This article has been accepted for inclusion in a future issue of this journal. Content is final as presented, with the exception of pagination.

# Deep Semantic Multimodal Hashing Network for Scalable Image-Text and Video-Text Retrievals

Lu Jin, Zechao Li, *Member, IEEE*, and Jinhui Tang, *Senior Member, IEEE*

***Abstract*— Hashing has been widely applied to multimodal retrieval on large-scale multimedia data due to its efficiency in computation and storage. In this article, we propose a novel deep semantic multimodal hashing network (DSMHN) for scalable image-text and video-text retrieval. The proposed deep hashing framework leverages 2-D convolutional neural networks (CNN) as the backbone network to capture the spatial information for image-text retrieval, while the 3-D CNN as the backbone network to capture the spatial and temporal information for video-text retrieval. In the DSMHN, two sets of modality-specific hash functions are jointly learned by explicitly preserving both intermodality similarities and intramodality semantic labels. Specifically, with the assumption that the learned hash codes should be optimal for the classification task, two stream networks are jointly trained to learn the hash functions by embedding the semantic labels on the resultant hash codes. Moreover, a unified deep multimodal hashing framework is proposed to learn compact and high-quality hash codes by exploiting the feature representation learning, intermodality similarity-preserving learning, semantic label-preserving learning, and hash function learning with different types of loss functions simultaneously. The proposed DSMHN method is a generic and scalable deep hashing framework for both image-text and video-text retrievals, which can be flexibly integrated with different types of loss functions. We conduct extensive experiments for both single-modal- and cross-modal-retrieval tasks on four widely used multimodal-retrieval data sets. Experimental results on both image-text- and video-text-retrieval tasks demonstrate that the DSMHN significantly outperforms the state-of-the-art methods.**

## I. Introduction

With the rapid advancement in social network, recent years have witnessed the explosive growth of social multimedia data including images, user-provided textual captions, and videos. When searching the topics of interests, it is desirable to return relevant results such as images, textual contents, and videos timely and accurately for the users [1]. As there is a huge semantic gap among different modalities, it is challenging when searching relevant multimodal contents among a great amount of heterogenous multimedia data. Consequently, it is imperative to develop efficient methods to support fast and effective similarity search for large-scale heterogenous multimedia data.

Hashing [2]–[7] has received a great deal of attention recently in the field of computer vision [8]–[10] and multimedia retrieval [11], [12]. The core idea of hashing is to project high-dimensional data points into compact binary hash codes, while preserving the similarity between the data points in the Hamming space. Compared with the real-valued features, hash codes greatly reduce the storage cost and allow fast Hamming distance computation with the XOR operation, resulting in efficient similarity search on large-scale databases. Existing multimodal hashing methods [13], [14] can be roughly divided into single-modal hashing methods and cross-modal hashing methods. Given a query (i.e., a query image), single-modal hashing methods focus on searching for the relevant contents among the same modal-retrieval database. Well-known single-modal hashing methods include locality-sensitive hashing (LSH) [15], iterative quantization (ITQ) [16], deep supervised hashing (DSH) [17], and supervised semantics-preserving deep hashing (SSDH) [18]. Multimodal data are usually represented with distinct feature representations in heterogenous feature spaces, which makes searching across multimodal data (i.e., image, text, and video) particularly challenging. Most of the cross-modal hashing methods [19]–[21] attempt to map heterogenous features to a common space so that the correlation structures of different modalities can be explored. However, very little efforts have been made to address both of the single-modal and cross-modal retrievals, which really limits the real application for multimedia retrieval.

Due to its great potential in learning powerful feature representation, deep hashing methods [22]–[26] have shown superior performance to the traditional hashing methods that adopt the handcraft features to learn the hash functions. By jointly learning the feature representation and hash functions, the feature representation is optimally compatible with hashing learning, thus leading to discriminative and high-quality binary hash codes. For most deep single-modal hashing methods [27]–[29], the foremost concern is to approximate the hash functions by preserving the intramodality similarity, which is insufficient to build the correlation structure of different modalities. Meanwhile, although latest deep cross-modal hashing methods [30]–[32] can well preserve the intermodality

Manuscript received September 1, 2019; revised March 5, 2020; accepted May 15, 2020. This work was supported in part by the National Key Research and Development Program of China under Grant 2018AAA0102002, in part by the National Natural Science Foundation of China under Grant 61720106004 and Grant 61732007, and in part by the Natural Science Foundation of Jiangsu Province under Grant BK20170033. *(Corresponding author: Zechao Li.)*

The authors are with the School of Computer Science and Engineering, Nanjing University of Science and Technology, Nanjing 210094, China (e-mail: lujin505@gmail.com; zechao.li@njust.edu.cn; jinhuitang@njust.edu.cn).

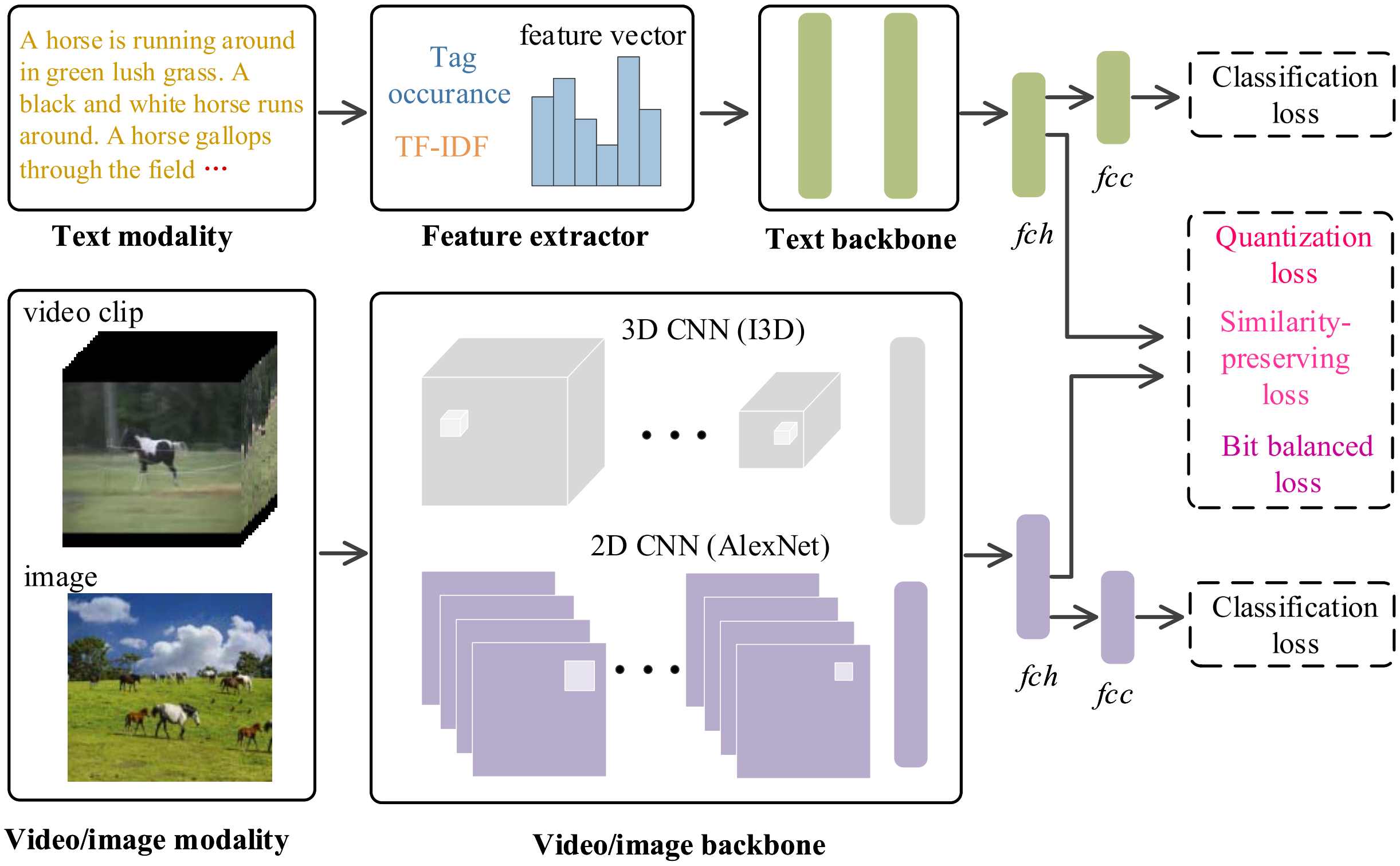

Fig. 1. Overview of the proposed DSMHN. The network learns the binary hash codes by exploring the spatial information (and temporal information) of images (videos) with four constraints on the hash layer ($fch$) of the network. First, the intermodality similarity-preserving loss is minimized to make the hash codes of different modalities consistent. Second, the classification loss is used to minimize the prediction errors of the learned hash codes. Third, the bit-balanced loss is minimized to make the hash bits evenly distributed. Finally, the quantization loss forces the outputs of the hash layer to be close to $-1$ and 1. Best viewed in color version.

correlation structure of multiple modalities, these methods are limited in single-modal retrieval because of lacking in preserving the intramodality semantic label information for hashing learning. In order to address these limitations, we propose a generalized deep hashing framework for scalable yet efficient image-text and video-text retrievals.

To this end, we propose a novel deep semantic multimodal hashing network (DSMHN) for scalable single-modal and cross-modal retrievals, which is shown in Fig. 1. To learn efficient modality-specific hash functions, the intermodality similarity and the intramodality semantic information are jointly exploited to guide the learning process of hash functions. Moreover, the learned hash codes are expected to be optimally compatible with the label predicting to capture effectively both intermodality correlation and intramodality semantic information. Based on this, the objective of the DSMHN is to minimize the pairwise similarity-preserving loss and the prediction errors of the learned hash codes simultaneously. The DSMHN can be flexibly trained with different types of loss functions under a generalized deep hashing framework. The proposed framework adopts 2-D convolutional neural networks (CNN) as the backbone network to capture the spatial information for image-text retrieval, while the 3-D CNN as the backbone network to capture the spatial and temporal information for video-text retrieval. Extensive experimental results show the superior performance of the proposed hashing framework to the representative hashing methods for image-text retrieval and video-text retrieval.

The key contributions of the proposed method are summarized as follows.

1) We propose a generalized deep multimodal hashing framework that explores feature representation learning, intermodality similarity preserving, intramodality semantic label preserving, and hash function learning with different types of loss functions simultaneously for scalable image-text and video-text retrievals.
2) The proposed model learns binary hash codes by directly embedding the semantic labels on the similarity-preserving hash codes and can be flexibly trained with different types of pairwise similarity-preserving loss functions.
3) The proposed model can well explore the spatial information of the images for image-text retrieval, while the spatial and temporal information for video-text retrieval.

The rest of this article is organized as follows. We briefly review the related works in Section II. The proposed method and the optimization algorithm are elaborated in Section III. In Section IV, experiments are conducted and analyzed followed by the conclusions in Section V.

## II. Related Works

As mentioned above, existing multimodal hashing methods can be roughly divided into two categories: single-modal hashing methods and cross-modal hashing methods. We briefly review these two kinds of methods in the following.

Single-modal hashing methods learn the hash functions by preserving the intramodality similarity that is defined by using the semantic label or the feature representation in the original feature space. Most earlier works, such as spectral hashing (SH) [3], ITQ [16], binary reconstructive embedding (BRE) [33], anchor graph hashing (AGH) [34], and neighborhood discriminant hashing (NDH) [2], explore the underlying data structure in the original feature space to learn the hash functions. $K$-nearest neighbor hashing (KNNH) [35] proposes conditional entropy minimization to make the data point and its k-nearest neighbors share the same hash codes. DistillHash [36] exploits the local structure to collect a distilled data set, which is further used to learn the hash codes under the Bayesian learning framework. Another successful attempt has been made to leverage the supervised label information to improve the quality of the binary hash codes [37]. Fast supervised discrete hashing (FSDH) [38] directly learns label-preserving binary hash codes under the discrete constraint. By generating the list of ranking orders from the word embeddings of the semantic labels, discrete semantic ranking hashing (DSeRH) [39] attempts to encode the semantic rank orders to learn the binary hash codes with the triplet ranking loss. Furthermore, asymmetric multivalued hashing (AMVH) [40], asymmetric binary coding (ASH) [41], and asymmetric DSH (ADSH) [42] use the asymmetric similarity metric to learn two different sets of hash functions for image retrieval. References [43] and [44] use the generative adversarial network to augment data for hash learning.

Cross-modal hashing methods learn modality-specific hash functions by projecting hierarchical feature representations into a common space. Cross-view hashing (CVH) [45], coregularized hashing [46], and intermedia hashing (IMH) [47] use the unlabeled training data to generate binary hash codes. As the supervised information is not exploited, these methods cannot produce desirable hash codes. Some recent works use the labeled and unlabeled training data to learn more efficient hash functions. Cross-modality similarity-sensitive hashing (CMSSH) [48] adopts the boosting algorithm to solve efficiently the cross-modal similarity learning by projecting the input data from two different spaces into the Hamming space. Collective matrix factorization hashing (CMFH) [49] uses collective matrix factorization to learn the cross-view hash functions by exploiting the latent structure of different modalities. Similarly, latent semantic sparse hashing (LSSH) [50] employs the matrix factorization and sparse coding to learn the latent concepts and the salient structure from the text and image, respectively. Semantic topic multimodal hashing (STMH) [51] learns a common subspace by capturing multiple semantic topics from multimedia data and then generates binary hash codes by checking the appearance of the semantic topics. Semantic correlation maximization (SCM) [13] extends the canonical correlation analysis in the supervised manner. Semantics-preserving hashing (SePH) [52] learns the joint binary hash codes by minimizing the Kullback–Leibler divergence between the hash codes and the semantic affinities, and then learns the cross-view hash functions with the kernel logistic regression. Linear subspace ranking hashing (LSRH) [53] learns the ranking-based hash functions by exploiting the ranking structure of the feature space.

Recently, CNN [54]–[56] have shown an impressive learning potential for many vision tasks such as image classification [57]–[60], image segmentation [61], [62], image captioning [63], [64], and multimedia retrieval [65]–[67]. The success of CNNs on those vision tasks indicates that the learned hierarchical representation can well preserve the underlying semantic structure of the input data from different modalities. CNNH [68] is a two-step hashing method that learns the hash codes and the deep hash functions separately for image retrieval. In [29], [69], and [70], the triplet ranking loss is designed to learn the hash functions with the CNN model by capturing the triplet-based relative similarity relationships of the image pairs. Multimodal similarity-preserving hashing (MMNN) [71] employs a siamese network to learn the cross-view hash functions by jointly preserving the intermodality and intramodality similarities of different modalities. The correlation hashing network (CHN) [72] exploits the multimodal correlation structure of the multimedia data under a deep network architecture. DCMH [73] directly learns unified discrete hash codes under the binary constraint. Cross-modal Hamming hashing (CMHH) [74] learns efficient binary hash codes by introducing the focal loss to make the Hamming distance of similar cross-modal pairs larger than the Hamming radius threshold. Attention-aware deep adversarial hashing [75] proposes an adversarial hashing framework based on the attention mechanism to explore the useful cross-modal information. Li *et al.* [76] presented a cross-modal hash method by incorporating the adversarial learning to maximize the semantic correlation and representation consistency between different modalities in the self-supervised manner.

Different from the aforementioned methods, we propose a generic deep hashing framework for both single-modal and cross-modal-retrieval tasks by integrating the feature representation learning, the intermodality similarity learning, the intramodality semantic label learning, and the hashing learning into one unified framework. It can well explore the spatial information of the images based on the 2-D CNN for image-text retrieval, while well capture the spatial and temporal information based on the 3-D CNN for video-text retrieval.

## III. Deep Semantic Multimodal Hashing Network

### A. Mathematical Notations and Problem Definition

Throughout this article, uppercase bold font characters are used to denote the matrices and lowercase bold font characters are used to denote the vectors. Let $T_{\mathcal{X}} = \{\mathrm{x_i}\}_{i=1}^N \in \mathbb{R}^{N \times d_{\mathcal{X}}}$ denote a set of $d_{\mathcal{X}}$-dimensional data points from the image/video modality $\mathcal{X}$. Similarly, we use $T_{\mathcal{Y}} = \{\mathrm{y_i}\}_{i=1}^N \in \mathbb{R}^{N \times d_{\mathcal{Y}}}$ to define a set of $d_{\mathcal{Y}}$-dimensional data points from the text modality $\mathcal{Y}$. We assume each instance (i.e., $x_i$ or $y_i$) belongs to one or more categories, and its corresponding label denotes $\mathbf{g} \in \{0, 1\}^C$, where $C$ is the number of the categories and the nonzero elements in $\mathbf{g}$ indicate that the instance is associated with the corresponding categories.

We target to learn the $L$-bit binary codes $\mathbf{B}_* \in \{-1, 1\}^{L \times N}$ for each modality as well as two sets of modality-specific hash functions $\mathcal{H}_* = [h_*^1, \ldots, h_*^L]$ with deep networks, where $* \in \{\mathcal{X}, \mathcal{Y}\}$ is a placeholder and $h_*^l$ is used to generate the $l$th binary code $b_*^l \in \{-1, 1\}$. After we obtain the modality-specific hash functions, new coming samples from different modalities can be easily projected to the common Hamming space. The proposed hashing framework aims to learn jointly the modality-specific hash functions and hierarchical representations in an end-to-end manner by exploring the intermodality semantic correlation and the semantic information simultaneously. In addition, each hash bit of the binary hash codes is expected to be zero-mean over the training data. In the following, we will describe the proposed deep hashing framework in detail.

### B. Deep Modality-Specific Hash Functions

Deep learning techniques have shown superiority in many computer vision tasks such as classification, segmentation, and image retrieval. CNNs especially have achieved great success in various visual applications due to the effectiveness in capturing high-level semantic structure from raw data. In contrast to the handcrafted features, the CNN can automatically learn the hierarchical feature representation that is invariant to the irrelevant variations such as translation and rotation. In this article, we propose to learn the modality-specific hash functions and feature representations synchronously for multimodal data by employing the deep networks. To capture the spatial information of the images, a 2-D CNN is adopted as the backbone network, while a 3-D CNN [77] is used as the backbone network to capture the spatial and temporal information of the videos.

Assume the deep network has $M_\mathcal{X}$ and $M_\mathcal{Y}$ layers for the image/video modality $\mathcal{X}$ and the text modality $\mathcal{Y}$, respectively. For the $m$th layer of the deep network, we can calculate the outputs of this layer as follows:

$$\mathbf{Z}_*^m = \psi\left(\mathbf{W}_*^m \mathbf{Z}_*^{m-1} + \mathbf{c}_*^m\right), \quad \text{for } m = 1, \ldots, M_* \tag{1}$$

where $\psi(\cdot)$ is the nonlinear activation function, $\mathbf{W}_*^m \in \mathbb{R}^{d_m^* \times d_{m-1}^*}$ is the weight matrix for the $m$th layer, $\mathbf{c}_*^m \in \mathbb{R}^{d_m^*}$ is the bias vector, and $\mathbf{Z}_*^{m-1} \in \mathbb{R}^{d_{m-1}^* \times N}$ is the outputs of the $(m-1)$th layer. In order to learn the binary hash codes, the hash layer [i.e., the $(M_* - 1)$th layer] of the deep network is used to construct the modality-specific hash functions. Specifically, the modality-specific binary codes can be computed by

$$\mathbf{B}^* = \text{sign}\left(\mathbf{Z}_*^{M_*-1}\right) \tag{2}$$

where $\text{sign}(a) = 1$ if $a \geq 0$; otherwise, $\text{sign}(a) = -1$, and $\mathbf{Z}_*^{M_*-1}$ is the output of the $(M_* - 1)$th layer.

### C. Intermodality Similarity-Preserving Binary Codes

To capture the correlation of different modalities, the intermodality similarity is preserved during hash function learning. First, we construct the intermodality similarity $\mathbf{S}$ as

$$s_{ij} = \begin{cases} 1, & \text{if } \mathbf{g}_i^T \mathbf{g}_j > 0 \\ -1, & \text{otherwise} \end{cases} \tag{3}$$

where $\mathbf{g}_i$ and $\mathbf{g}_j$ are the label vectors for the $i$th and $j$th instances, respectively. For each cross-modal pair $(x_i, y_j, s_{ij})$, if $s_{ij} = 1$, the corresponding binary codes $\mathbf{b}_i^\mathcal{X}$ and $\mathbf{b}_j^\mathcal{Y}$ should be similar to each other; otherwise, $\mathbf{b}_i^\mathcal{X}$ and $\mathbf{b}_j^\mathcal{Y}$ should be dissimilar. In other words, if $s_{ij} = 1$, the similarity between $\mathbf{b}_i^\mathcal{X}$ and $\mathbf{b}_j^\mathcal{Y}$ should be close to 1; otherwise, it should be close to $-1$. Specifically, the widely used code inner product is used to measure the similarity of the two data points. Therefore, the cross-modal similarity between $\mathbf{b}_i^\mathcal{X}$ and $\mathbf{b}_j^\mathcal{Y}$ can be computed by

$$c_{ij} = \left(\mathbf{b}_i^\mathcal{X}\right)^T \mathbf{b}_j^\mathcal{Y} / L \tag{4}$$

where $L$ is the length of the hash codes.

In general, we can formulate the intermodality similarity-preserving learning problem as follows:

$$\min_{\{\mathbf{W}_\mathcal{X}^m, \mathbf{b}_\mathcal{X}^m\}, \{\mathbf{W}_\mathcal{Y}^m, \mathbf{b}_\mathcal{Y}^m\}} \sum_{s_{ij} \in \mathbf{S}} \ell\left(c_{ij}, s_{ij}\right), \quad \text{for } m = 1, \ldots, M_* - 1 \tag{5}$$

where $\ell(\cdot, \cdot)$ is the loss function to enforce $c_{ij}$ and $s_{ij}$ to be close and $\{\mathbf{W}_*^m, \mathbf{b}_*^m\}$ is the parameters of the $m$th layer. In (5), we can use any proper loss function, and several widely used loss functions are discussed as follows.

1) L1 Loss: The L1 loss minimizes the absolute differences between the estimated value $c_{ij}$ and the target value $s_{ij}$

$$\ell^{L1}\left(c_{ij}, s_{ij}\right) = |c_{ij} - s_{ij}| \tag{6}$$

where $|a|$ denotes the absolute value of $a$.

2) Euclidean Loss: The Euclidean loss is the L2-norm of the error between the estimated values and the target values

$$\ell^{L2}\left(c_{ij}, s_{ij}\right) = \frac{1}{2}\left(c_{ij} - s_{ij}\right)^2. \tag{7}$$

3) Hinge Loss: The hinge loss is widely used in the SVMs for maximum-margin classification, which is defined as

$$\ell^{hg}\left(c_{ij}, s_{ij}\right) = \begin{cases} \max\left(0, 0.5 - c_{ij}\right), & \text{if } s_{ij} = 1 \\ c_{ij}, & \text{if } s_{ij} = -1. \end{cases} \tag{8}$$

4) Contrastive Loss: The contrastive loss is defined as

$$\ell^{ct}\left(c_{ij}, s_{ij}\right) = \begin{cases} d_{ij}, & \text{if } s_{ij} = 1 \\ \max\left(0, 0.5 - d_{ij}\right), & \text{if } s_{ij} = -1 \end{cases} \tag{9}$$

where $d_{ij} = (1/L)\|\mathbf{b}_i^\mathcal{X} - \mathbf{b}_j^\mathcal{Y}\|_2^2 = 2(1 - c_{ij})$ denotes the distance between $\mathbf{b}_i^\mathcal{X}$ and $\mathbf{b}_j^\mathcal{Y}$.

### D. Label-Preserving Binary Codes

Although the intermodality similarity is leveraged to explore the correlation between different modalities, the intramodality label information is not fully exploited for cross-view retrieval. Recent works of image retrieval have indicated that high-quality hash functions can be learned when the semantic label is exploited to capture the high-level semantic information for hash learning. To exploit fully the semantic labels, the learned binary hash codes are directly used for label predicting. That is to say, the binary hash codes are expected to be optimal for the jointly learned classifier.

In addition to exploring the correlations between different modalities, we make full use of the semantic labels to learn the binary hash codes for the image/video and text modalities, respectively. In particular, the last layer (named the $fcc$ layer) of the modality-specific network is trained for the classification task. In this article, we adopt the widely used cross-entropy loss for classification, which can be written as

$$\begin{aligned}\mathcal{L}_c\left(\mathbf{B}^*; \mathbf{W}_*^{M_*}, \mathbf{c}_*^{M_*}\right) &= -\frac{1}{N}\sum_{i=1}^{N}\ell_c^*\left(\mathbf{b}_i^*; \mathbf{W}_*^{M_*}, \mathbf{c}_*^{M_*}\right)\\ &= -\frac{1}{N}\sum_{i=1}^{N}\sum_{c=1}^{C} g_{ic}\log \hat{g}_{ic}\\ &\quad - (1-g_{ic})\log(1-\hat{g}_{ic})\end{aligned} \tag{10}$$

where $\mathbf{W}_*^{M_*}$ and $\mathbf{c}_*^{M_*}$ define the weight matrix and bias vector for the $fcc$ layer of the network, $g_{ic}$ is the $c$th element of $\mathbf{g}_i$, $\hat{g}_{ic} = \phi(\mathbf{w}_{c*}^{M_*}\mathbf{b}_i^* + c_{c*}^{M_*})$ estimates the probability that the $i$th instance is classified as the $c$th category, $\mathbf{w}_{c*}^{M_*}$ is the $c$th row of $\mathbf{W}_*^{M_*}$, $c_{c*}^{M_*}$ is the $c$th element of $\mathbf{c}_*^{M_*}$, and $\phi(a) = (1+\exp(-a))^{-1}$ defines the sigmoid function.

### *E. Overall Objective and Optimization*

As mentioned above, we propose to learn the hash functions with the deep neural networks by simultaneously exploring the intramodality semantic structure and the intermodality correlations between different modalities. By combining (5) and (10), the proposed framework is formulated as

$$\begin{aligned}&\min_{\Phi_\mathcal{X},\Phi_\mathcal{Y}} \mathcal{O} = \sum_{s_{ij}\in\mathbf{S}}\ell_{ij} + \alpha\left(\mathcal{L}_c^\mathcal{X} + \mathcal{L}_c^\mathcal{Y}\right)\\ &\text{s.t. } \mathbf{B}^\mathcal{X},\quad \mathbf{B}^\mathcal{Y} \in \{-1,1\}^{L\times N}\end{aligned} \tag{11}$$

where $\ell_{ij}$ stands for $\ell(c_{ij}, s_{ij})$, $\mathcal{L}_c^\mathcal{X}$ and $\mathcal{L}_c^\mathcal{Y}$ are short for $\mathcal{L}_c(\mathbf{B}^\mathcal{X}; \mathbf{W}_\mathcal{X}^{M_\mathcal{X}}, \mathbf{c}_\mathcal{X}^{M_\mathcal{X}})$ and $\mathcal{L}_c(\mathbf{B}^\mathcal{Y}; \mathbf{W}_\mathcal{Y}^{M_\mathcal{Y}}, \mathbf{c}_\mathcal{Y}^{M_\mathcal{Y}})$, respectively, $\Phi_\mathcal{X} = \{\mathbf{W}_\mathcal{X}^m, \mathbf{c}_\mathcal{X}^m\}_{m=1}^{M_\mathcal{X}}$ is the parameter set of the image/video network, $\Phi_\mathcal{Y} = \{\mathbf{W}_\mathcal{Y}^{m'}, \mathbf{c}_\mathcal{Y}^{m'}\}_{m'=1}^{M_\mathcal{Y}}$ is the parameter set of the text network, and $\alpha > 0$ is a hyperparameter.

To make the hash codes balanced, most hashing methods enforce each bit of the binary codes to be mean-zero over the training data [18]. Similarly, we impose the bit-balance constraints under the proposed deep hashing framework; therefore, problem (11) can be reformulated as

$$\begin{aligned}&\min_{\Phi_\mathcal{X},\Phi_\mathcal{Y}} \mathcal{O} = \sum_{s_{ij}\in\mathbf{S}}\ell_{ij} + \alpha\left(L_c^\mathcal{X} + L_c^\mathcal{Y}\right)\\ &\text{s.t. } \mathbf{B}^\mathcal{X},\quad \mathbf{B}^\mathcal{Y} \in \{-1,1\}^{L\times N},\\ &\qquad \mathbf{B}^\mathcal{X}\mathbf{1} = 0,\quad \mathbf{B}^\mathcal{Y}\mathbf{1} = 0\end{aligned} \tag{12}$$

where $\mathbf{1} \in \mathbb{R}^N$ is a vector with all the elements being 1.

The optimization problem in (12) is NP-hard because of the discrete constraints on $\mathbf{B}^\mathcal{X}$ and $\mathbf{B}^\mathcal{Y}$. To make it tractable, we relax $\mathbf{B}^\mathcal{X}$ and $\mathbf{B}^\mathcal{Y}$ to be continuous values, and then, (2) can be reformulated as

$$\mathbf{B}^* = \mathbf{Z}_*^{M_*-1} \tag{13}$$

where $\mathbf{Z}_*^{M_*-1}$ is the output of the hash layer ($fch$ layer). Furthermore, the optimization problem in (12) becomes

$$\begin{aligned}&\min_{\Phi_\mathcal{X},\Phi_\mathcal{Y}} \mathcal{O} = \sum_{s_{ij}\in\mathbf{S}}\ell_{ij} + \alpha\left(L_c^\mathcal{X} + L_c^\mathcal{Y}\right)\\ &\text{s.t. } \mathbf{Z}_\mathcal{X}^{M_\mathcal{X}-1},\quad \mathbf{Z}_\mathcal{Y}^{M_\mathcal{Y}-1} \in \{-1,1\}^{L\times N},\\ &\qquad \mathbf{Z}_\mathcal{X}^{M_\mathcal{X}-1}\mathbf{1} = 0,\quad \mathbf{Z}_\mathcal{Y}^{M_\mathcal{Y}-1}\mathbf{1} = 0.\end{aligned} \tag{14}$$

To make the hash codes be $-1$ or 1, we introduce a quantization error term as follows:

$$\mathcal{L}_q = \frac{1}{2N}\left(\||\mathbf{Z}_\mathcal{X}^{M_\mathcal{X}-1}| - \mathbf{1}_{L\times N}\|_F^2 + \||\mathbf{Z}_\mathcal{Y}^{M_\mathcal{Y}-1}| - \mathbf{1}_{L\times N}\|_F^2\right). \tag{15}$$

By minimizing the quantization error term $\mathcal{L}_q$, the entries of $\mathbf{Z}_\mathcal{X}^{M_\mathcal{X}-1}$ and $\mathbf{Z}_\mathcal{Y}^{M_\mathcal{Y}-1}$ are enforced to be close to $-1$ and 1. As a result, the optimization problem in (14) can be reformulated as

$$\min_{\Phi_\mathcal{X},\Phi_\mathcal{Y}} \mathcal{O} = \sum_{s_{ij}\in\mathbf{S}}\ell_{ij} + \alpha\left(\mathcal{L}_c^\mathcal{X} + \mathcal{L}_c^\mathcal{Y}\right) + \beta\mathcal{L}_q + \gamma\mathcal{L}_b \tag{16}$$

where $\mathcal{L}_b = (1/2N)(\|\mathbf{Z}_\mathcal{X}^{M_\mathcal{X}-1}\mathbf{1}_N\|_2^2 + \|\mathbf{Z}_\mathcal{Y}^{M_\mathcal{Y}-1}\mathbf{1}_N\|_2^2)$ defines the bit-balance penalty term and $\beta, \gamma > 0$ are the two hyperparameters. Note that, with sufficiently large $\gamma$, the binary hash codes $\mathbf{B}^*$ would be balanced as much as possible.

In order to solve the above problem, we adopt the widely used stochastic gradient descent (SGD) with minibatch to train the whole deep networks. Specifically, the backpropagation (BP) strategy is employed to update the parameters of each layer. In general, the joint optimization problem in (16) is nonconvex with respect to $\Phi_\mathcal{X}$ and $\Phi_\mathcal{Y}$. In this article, we propose to learn alternatively $\Phi_\mathcal{X}$ and $\Phi_\mathcal{Y}$, where one parameter set is optimized while with another one fixed. In the following, we describe the optimization algorithm in detail.

*Update $\Phi_\mathcal{X}$:* When $\Phi_\mathcal{Y}$ is fixed, for the $m$th layer of the image/video network, the gradients of the overall objective function $\mathcal{O}$ with respect to the parameters $\mathbf{W}_\mathcal{X}^m$ and $\mathbf{b}_\mathcal{X}^m$ can be derived by

$$\begin{aligned}\frac{\partial\mathcal{O}}{\partial\mathbf{W}_\mathcal{X}^m} &= \xi_\mathcal{X}^m\left(\mathbf{z}_{\mathcal{X}i}^{m-1}\right)^T\\ \frac{\partial\mathcal{O}}{\partial\mathbf{c}_\mathcal{X}^m} &= \xi_\mathcal{X}^m\end{aligned} \tag{17}$$

where $\mathbf{z}_{\mathcal{X}i}^{m-1}$ is the $i$th column of $\mathbf{Z}_\mathcal{X}^{m-1}$. For the hidden layer (i.e., $m = 1, \ldots, M_\mathcal{X} - 2$), $\xi_\mathcal{X}^m$ can be derived as follows:

$$\xi_\mathcal{X}^m = \left(\left(\mathbf{W}_\mathcal{X}^{m+1}\right)^T\xi_\mathcal{X}^{m+1}\right)\odot\psi'\left(\mathbf{W}_\mathcal{X}^m\mathbf{z}_{\mathcal{X}i}^{m-1} + \mathbf{c}_\mathcal{X}^m\right) \tag{18}$$

where $\odot$ denotes the elementwise multiplication and $\psi'(\cdot)$ defines the gradient of $\psi(\cdot)$.

For the classification layer (i.e., $m = M_\mathcal{X}$), $\xi_\mathcal{X}^{M_\mathcal{X}}$ can be computed by

$$\xi_\mathcal{X}^{M_\mathcal{X}} = \frac{\alpha}{N}\left(\hat{\mathbf{g}}_i^\mathcal{X} - \mathbf{g}_i\right)\odot\psi'\left(\mathbf{W}_\mathcal{X}^{M_\mathcal{X}}\mathbf{z}_{\mathcal{X}i}^{M_\mathcal{X}-1} + \mathbf{c}_\mathcal{X}^{M_\mathcal{X}}\right) \tag{19}$$

where $\hat{\mathbf{g}}_i^\mathcal{X} = \phi(\mathbf{W}_\mathcal{X}^{M_\mathcal{X}}\mathbf{z}_{\mathcal{X}i}^{M_\mathcal{X}-1} + \mathbf{c}_\mathcal{X}^{M_\mathcal{X}})$.

For the hash layer (i.e., $m = M_{\mathcal{X}} - 1$), $\xi_{\mathcal{X}}^{M_{\mathcal{X}}-1}$ can be calculated by

$$\xi_{\mathcal{X}}^{M_{\mathcal{X}}-1} = \left(\frac{\partial \ell_{ij}}{\partial \mathbf{z}_{\mathcal{X}}^{M_{\mathcal{X}}-1}} + \beta \frac{\partial \mathcal{L}_q^{\mathcal{X}}}{\partial \mathbf{z}_{\mathcal{X}}^{M_{\mathcal{X}}-1}} + \gamma \frac{\partial \mathcal{L}_b^{\mathcal{X}}}{\partial \mathbf{z}_{\mathcal{X}}^{M_{\mathcal{X}}-1}}\right) \odot \psi'\left(\mathbf{W}_{\mathcal{X}}^{M_{\mathcal{X}}-1}\mathbf{z}_{\mathcal{X}i}^{M_{\mathcal{X}}-2} + \mathbf{c}_{\mathcal{X}}^{M_{\mathcal{X}}-1}\right) \quad (20)$$

where

$$\frac{\partial \mathcal{L}_q^{\mathcal{X}}}{\partial \mathbf{z}_{\mathcal{X}}^{M_{\mathcal{X}}-1}} = \frac{1}{N}\left(|\mathbf{z}_{\mathcal{X}i}^{M_{\mathcal{X}}-1}| - \mathbf{1}\right) \odot \operatorname{sign}\left(\mathbf{z}_{\mathcal{X}i}^{M_{\mathcal{X}}-1}\right); \quad (21)$$

$$\frac{\partial \mathcal{L}_b^{\mathcal{X}}}{\partial \mathbf{z}_{\mathcal{X}}^{M_{\mathcal{X}}-1}} = \frac{1}{N}\mathbf{z}_{\mathcal{X}i}^{M_{\mathcal{X}}-1}\mathbf{1}_N. \quad (22)$$

Since different loss functions are used in the proposed deep hashing framework, the subgradient ($\partial \ell_{ij}/\partial \mathbf{Z}_{\mathcal{X}}^{M_{\mathcal{X}}-1}$) of different loss functions is discussed as follows:

$$\begin{cases} \frac{\partial \ell_{ij}^{L1}}{\partial \mathbf{z}_{\mathcal{X}}^{M_{\mathcal{X}}-1}} = \frac{1}{L}\operatorname{sign}(c_{ij} - s_{ij})\mathbf{z}_{\mathcal{Y}j}^{M_{\mathcal{Y}}-1} \\ \frac{\partial \ell_{ij}^{L2}}{\partial \mathbf{z}_{\mathcal{X}}^{M_{\mathcal{X}}-1}} = \frac{1}{L}(c_{ij} - s_{ij})\mathbf{z}_{\mathcal{Y}j}^{M_{\mathcal{Y}}-1} \\ \frac{\partial \ell_{ij}^{hg}}{\partial \mathbf{z}_{\mathcal{X}}^{M_{\mathcal{X}}-1}} = \frac{1}{L}\left((1 - s'_{ij}) - s'_{ij}I(\mu^{hg} > 0)\right)\mathbf{z}_{\mathcal{Y}j}^{M_{\mathcal{Y}}-1} \\ \frac{\partial \ell_{ij}^{ct}}{\partial \mathbf{z}_{\mathcal{X}}^{M_{\mathcal{X}}-1}} = \frac{2}{L}\left(s'_{ij} - (1 - s'_{ij})I(\mu^{ct} > 0)\right)\delta \end{cases} \quad (23)$$

where $s'_{ij} = \frac{1}{2}(s_{ij} + 1)$, $\mu^{hg} = 0.5 - c_{ij}$, $\mu^{ct} = 0.5 - d_{ij}$, $\delta = \mathbf{z}_{\mathcal{X}i}^{M_{\mathcal{X}}-1} - \mathbf{z}_{\mathcal{Y}j}^{M_{\mathcal{Y}}-1}$, and $I(condition)$ defines the indicator function that is equal to 1 if the $condition$ is true and 0 otherwise.

*Update $\Phi_{\mathcal{Y}}$:* With $\Phi_{\mathcal{X}}$ fixed, the gradients of the overall objective function $\mathcal{O}$ with respect to the parameters $\mathbf{W}_{\mathcal{Y}}^m$ and $\mathbf{b}_{\mathcal{Y}}^m$ of the $m$th layer for the text network are calculated by

$$\begin{aligned} \frac{\partial \mathcal{O}}{\partial \mathbf{W}_{\mathcal{Y}}^m} &= \xi_{\mathcal{Y}}^m\left(\mathbf{z}_{\mathcal{Y}i}^{m-1}\right)^T \\ \frac{\partial \mathcal{O}}{\partial \mathbf{c}_{\mathcal{Y}}^m} &= \xi_{\mathcal{Y}}^m \end{aligned} \quad (24)$$

where $\mathbf{z}_{\mathcal{Y}i}^{m-1}$ is the $i$th column of $\mathbf{Z}_{\mathcal{Y}}^{m-1}$.

If $m = 1, \ldots, M_{\mathcal{X}} - 2$, we can obtain $\xi_{\mathcal{Y}}^m$ by

$$\xi_{\mathcal{Y}}^m = \left(\left(\mathbf{W}_{\mathcal{Y}}^{m+1}\right)^T \xi_{\mathcal{Y}}^{m+1}\right) \odot \psi'\left(\mathbf{W}_{\mathcal{Y}}^m \mathbf{z}_{\mathcal{Y}i}^{m-1} + \mathbf{c}_{\mathcal{Y}}^m\right). \quad (25)$$

Otherwise, for the classification layer, $\xi_{\mathcal{Y}}^{M_{\mathcal{Y}}}$ can be computed by

$$\xi_{\mathcal{Y}}^{M_{\mathcal{Y}}} = \frac{\alpha}{N}\left(\hat{\mathbf{g}}_j^{\mathcal{Y}} - \mathbf{g}_j\right) \odot \psi'\left(\mathbf{W}_{\mathcal{Y}}^{M_{\mathcal{Y}}}\mathbf{z}_{\mathcal{Y}j}^{M_{\mathcal{Y}}-1} + \mathbf{c}_{\mathcal{Y}}^{M_{\mathcal{Y}}}\right) \quad (26)$$

where $\hat{\mathbf{g}}_j^{\mathcal{Y}} = \phi(\mathbf{W}_{\mathcal{Y}}^{M_{\mathcal{Y}}}\mathbf{z}_{\mathcal{Y}j}^{M_{\mathcal{Y}}-1} + \mathbf{c}_{\mathcal{Y}}^{M_{\mathcal{Y}}})$.

Similarly, for the hash layer, $\xi_{\mathcal{Y}}^{M_{\mathcal{Y}}-1}$ can be computed by

$$\xi_{\mathcal{Y}}^{M_{\mathcal{Y}}-1} = \left(\frac{\partial \ell_{ij}}{\partial \mathbf{z}_{\mathcal{Y}}^{M_{\mathcal{Y}}-1}} + \beta \frac{\partial \mathcal{L}_q^{\mathcal{Y}}}{\partial \mathbf{z}_{\mathcal{Y}}^{M_{\mathcal{Y}}-1}} + \gamma \frac{\partial \mathcal{L}_b^{\mathcal{Y}}}{\partial \mathbf{z}_{\mathcal{Y}}^{M_{\mathcal{Y}}-1}}\right) \odot \psi'\left(\mathbf{W}_{\mathcal{Y}}^{M_{\mathcal{Y}}-1}\mathbf{z}_{\mathcal{Y}i}^{M_{\mathcal{Y}}-2} + \mathbf{c}_{\mathcal{Y}}^{M_{\mathcal{Y}}-1}\right) \quad (27)$$

**Algorithm 1** Learning Algorithm for DSMHN

**Require:** Training set $T_{\mathcal{X}}$ and $T_{\mathcal{Y}}$.
**Ensure:** Network parameter sets $\Phi_{\mathcal{X}}$ and $\Phi_{\mathcal{Y}}$.
 **Initialization**
 Initialize Network parameter sets $\Phi_{\mathcal{X}}$ and $\Phi_{\mathcal{Y}}$, iteration number $N_{iter}$ and mini-batch size $N_{batch}$.
 **repeat:**
 **for** $iter = 1, \cdots, N_{iter}$ **do**
  Randomly select $N_{batch}$ cross-modal pairs from the training set $T_{\mathcal{X}}$ and $T_{\mathcal{Y}}$.
  **for** $m = 1, \ldots, M_*$ **do**
   For each selected cross-modal pair, compute $\mathbf{z}_*^m$ by forward propagation according to Eq. (1).
  **end for**
  **for** $m = M_*, \ldots, 1$ **do**
   Calculate the derivatives $\frac{\partial \mathcal{O}}{\partial \mathbf{W}_*^m}$ and $\frac{\partial \mathcal{O}}{\partial \mathbf{c}_*^m}$ according to Eq. (17) - (30).
   Update the parameters $\mathbf{W}_*^m$ and $\mathbf{c}_*^m$ using the BP algorithm.
  **end for**
 **end for**
 **until** a fixed number of iterations.

where

$$\frac{\partial \mathcal{L}_q^{\mathcal{Y}}}{\partial \mathbf{z}_{\mathcal{Y}}^{M_{\mathcal{Y}}-1}} = \frac{1}{N}\left(|\mathbf{z}_{\mathcal{Y}j}^{M_{\mathcal{Y}}-1}| - \mathbf{1}\right) \odot \operatorname{sign}\left(\mathbf{z}_{\mathcal{Y}j}^{M_{\mathcal{Y}}-1}\right), \quad (28)$$

$$\frac{\partial \mathcal{L}_b^{\mathcal{Y}}}{\partial \mathbf{z}_{\mathcal{Y}}^{M_{\mathcal{Y}}-1}} = \frac{1}{N}\mathbf{z}_{\mathcal{Y}j}^{M_{\mathcal{Y}}-1}\mathbf{1}_N, \quad (29)$$

$$\begin{cases} \frac{\partial \ell_{ij}^{L1}}{\partial \mathbf{z}_{\mathcal{Y}}^{M_{\mathcal{Y}}-1}} = \frac{1}{L}\operatorname{sign}(c_{ij} - s_{ij})\mathbf{z}_{\mathcal{X}i}^{M_{\mathcal{X}}-1} \\ \frac{\partial \ell_{ij}^{L2}}{\partial \mathbf{z}_{\mathcal{Y}}^{M_{\mathcal{Y}}-1}} = \frac{1}{L}(c_{ij} - s_{ij})\mathbf{z}_{\mathcal{X}i}^{M_{\mathcal{X}}-1} \\ \frac{\partial \ell_{ij}^{hg}}{\partial \mathbf{z}_{\mathcal{Y}}^{M_{\mathcal{Y}}-1}} = \frac{1}{L}\left((1 - s'_{ij}) - s'_{ij}I(\mu^{hg} > 0)\right)\mathbf{z}_{\mathcal{X}i}^{M_{\mathcal{X}}-1} \\ \frac{\partial \ell_{ij}^{ct}}{\partial \mathbf{z}_{\mathcal{Y}}^{M_{\mathcal{Y}}-1}} = \frac{-2}{L}\left(s'_{ij} - (1 - s'_{ij})I(\mu^{ct} > 0)\right)\delta. \end{cases} \quad (30)$$

Algorithm 1 summarizes the learning procedure of the proposed method.

## IV. Experiments

In this section, we conduct extensive experiments on four widely used multimodal data sets to validate the effectiveness of the proposed deep hashing method.

### A. Data Sets

Extensive experiments are conducted on three image-text data sets and one video-text data set. The image-text benchmark data sets include Wiki [78], MIRFlickr25k [79], and NUS-WIDE [80], and the Video-Text data set is MSR-VTT-10K [81].

The Wiki data set includes 2866 image-text pairs that are crawled from Wikipedia. Each image-text pair in this data set is annotated with one of the ten provided labels. We randomly select 20% of the image-text pairs as the query set, and the remaining pairs are used to form the database set, which is also used as the training set.

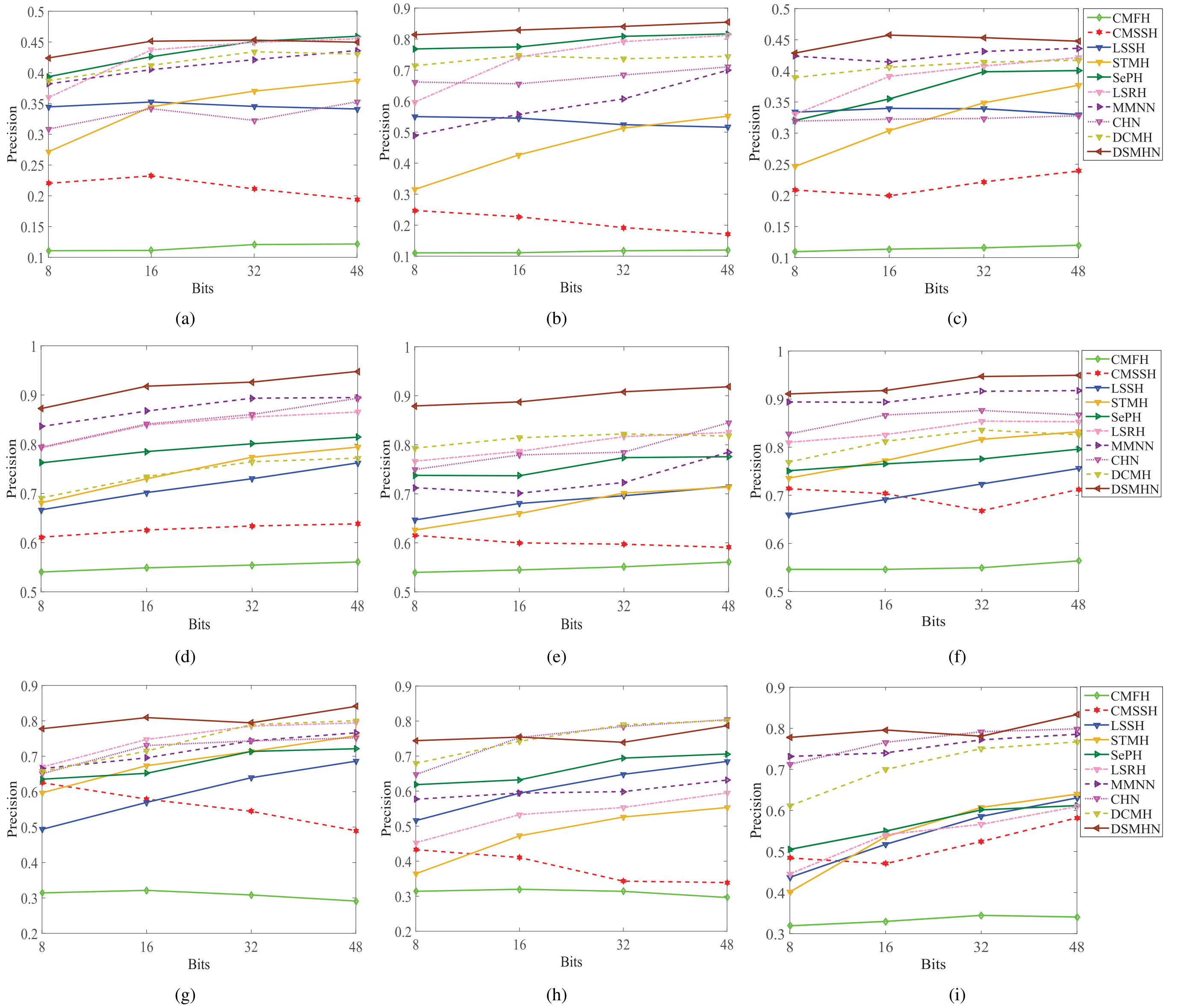

Fig. 2. Comparison in terms of P@100 on Wiki, MIRFlickr25k, and NUS-WIDE with 8, 16, 32, and 48 hash bits. (a) Wiki (image-query-text). (b) Wiki (text-query-image). (c) Wiki (image-query-image). (d) MIRFlickr25k (image-query-text). (e) MIRFlickr25k (text-query-image). (f) MIRFlickr25k (image-query-image). (g) NUS-WIDE (image-query-text). (h) NUS-WIDE (text-query-image). (i) NUS-WIDE (image-query-image).

The MIRFlickr25k data set consists of 25 000 images crawled from Flickr, in which each image is associated with its corresponding textual tags. The image-text pair in this data set is annotated with one or more of 24 provided labels. We keep the tags that appear at least 20 times, and the image-text pairs without any annotations and tags are removed. We randomly select 908 image-text pairs as the query set, and the remaining is used as the database set from which 5000 image-text pairs are randomly selected as the training set.

The NUS-WIDE data set contains 269 648 images collected from Flickr, where each image is associated with some textual tags. There are 81 ground-truth semantic concepts in this data set. We keep the top 21 most frequent concepts and get 195 834 image-text pairs for this data set. We randomly select 100 pairs per class as the query set, and the rest of the pairs are used as the database set, from which 500 pairs per class are selected to form the training set.

MSR-VTT-10K is the most recent large-scale benchmark data set for video-text retrieval and video captioning. It contains 10 000 Youtube video clips that belong to 20 categories. Each clip is annotated with 20 different text sentences. The data set is split into training set with 6513 clips and validation set with 2990 clips. We adopt the training set as the database for training and the validation set as the queries.

### *B. Compared Methods and Evaluation Metrics*

In order to evaluate the performance of the proposed deep hashing method, we compare it with several state-of-the-art methods, including six nondeep hashing methods (i.e., CMFH [49], LSSH [50], STMH [51], SCM [13], SePH [52], and LSRH [53]) and four deep hashing methods (i.e., MMNN [71], CHN [72], DCMH [73], and CMHH [74]). Those hashing methods have been introduced in Section II in detail. The source codes of those methods are available online expect for MMNN, CHN, DCMH, and CMHH that are carefully implemented by ourselves. The parameters of those methods are chosen according to the suggestions presented in the original papers. For the nondeep hashing methods, the images are represented by the 4096-D CNN features extracted from the last second fully connected layer of the AlexNet [54] network. The AlexNet network is used as the backbone for all the deep baselines for fair comparison. Furthermore, we extracted 1000-D and 2000-D TF-IDF features for the text in Wiki and MSR-VTT-10K, respectively. While for MIRFlickr25k and NUS-WIDE, the image tags are represented by the tag-occurrence feature vectors.

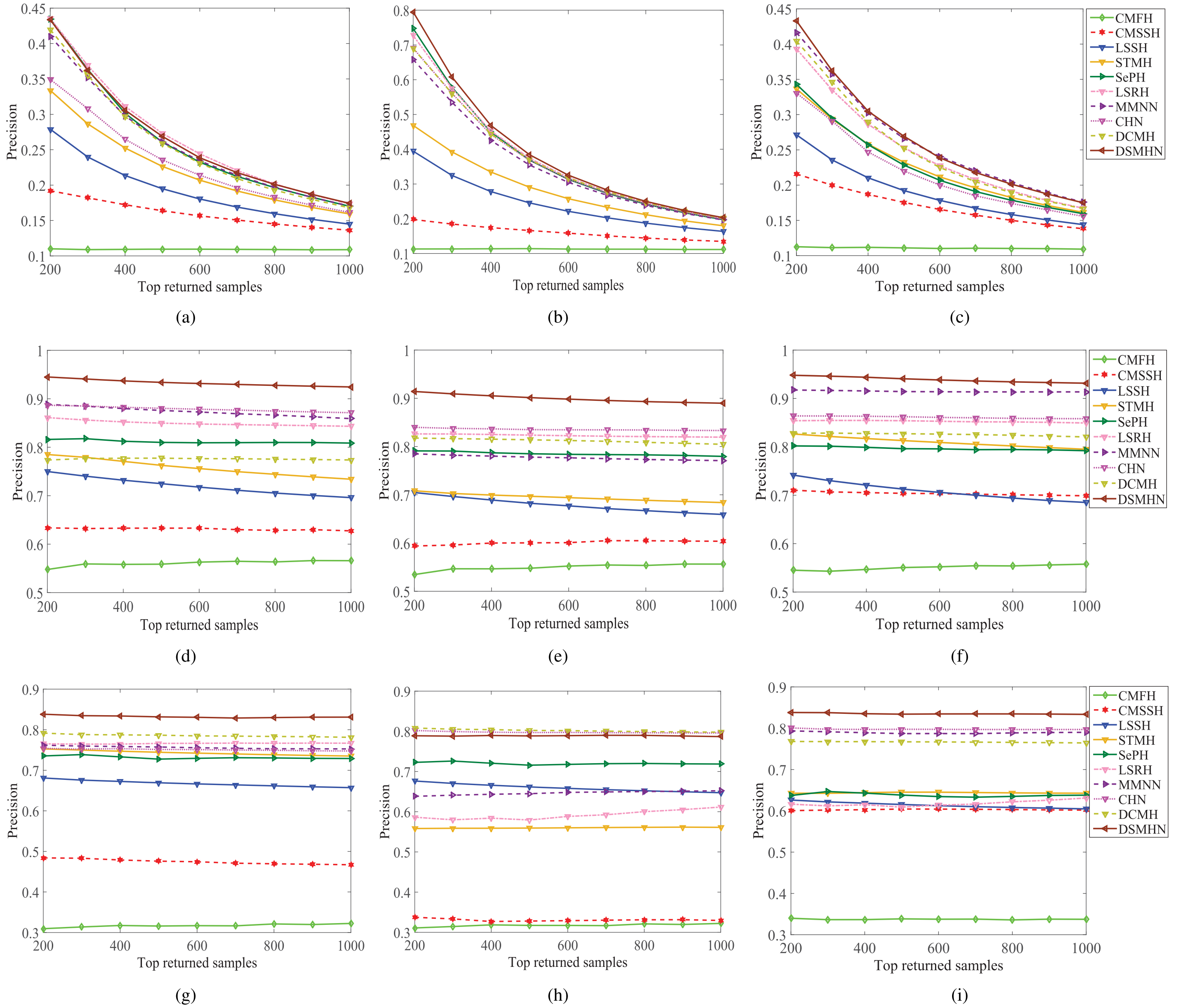

Fig. 3. Precision curves with different numbers of top returned samples for all the methods on Wiki, MIRFlickr25k, and NUS-WIDE with 48 hash bits. (a) Wiki (image-query-text). (b) Wiki (text-query-image). (c) Wiki (image-query-image). (d) MIRFlickr25k (image-query-text). (e) MIRFlickr25k (text-query-image). (f) MIRFlickr25k (image-query-image). (g) NUS-WIDE (image-query-text). (h) NUS-WIDE (text-query-image). (i) NUS-WIDE (image-query-image).

In this article, we evaluate the retrieval performance of image/video-query-text, text-query-image/video, and image/video-query-image/video. In addition, we evaluate the retrieval performance based on several widely used metrics including mean average precision (mAP), top-K precision (P@K), and precision-recall (PR).

### C. Experimental Setting

The proposed network contains two network streams: an image/video subnetwork and a text subnetwork. For the image/video subnetwork, we employ the AlexNet [54] and I3D [77] models as the image and video backbone, respectively. For hash learning, we add two fully connected layers at the top of the image/video backbone, as shown in Fig. 1. The first new fully connected layer (named the $fch$ layer) transforms the deep feature representation into compact hash codes. We set the number of the $fch$ layer as the length of the hash codes. To encourage the activations of the $fch$ layer binary, the $tanh$ activation function is used to make further the activations bounded between $-1$ and 1. To make the hash codes discriminative, the learned hash codes are expected to predict well the class labels. Specifically, the outputs of the $fch$ layer are fed to the second new fully connected layer (named the $fcc$ layer) for classification. The output number of the $fcc$ layer is the same as the categories. For the text subnetwork, it is comprised of four fully connected layers, where the first two layers construct the text backbone for feature representation learning and the last two layers are for hash learning. The output number of the first two layers are set to 4096 and the configuration of the last two layers is the same as the image/video subnetwork.

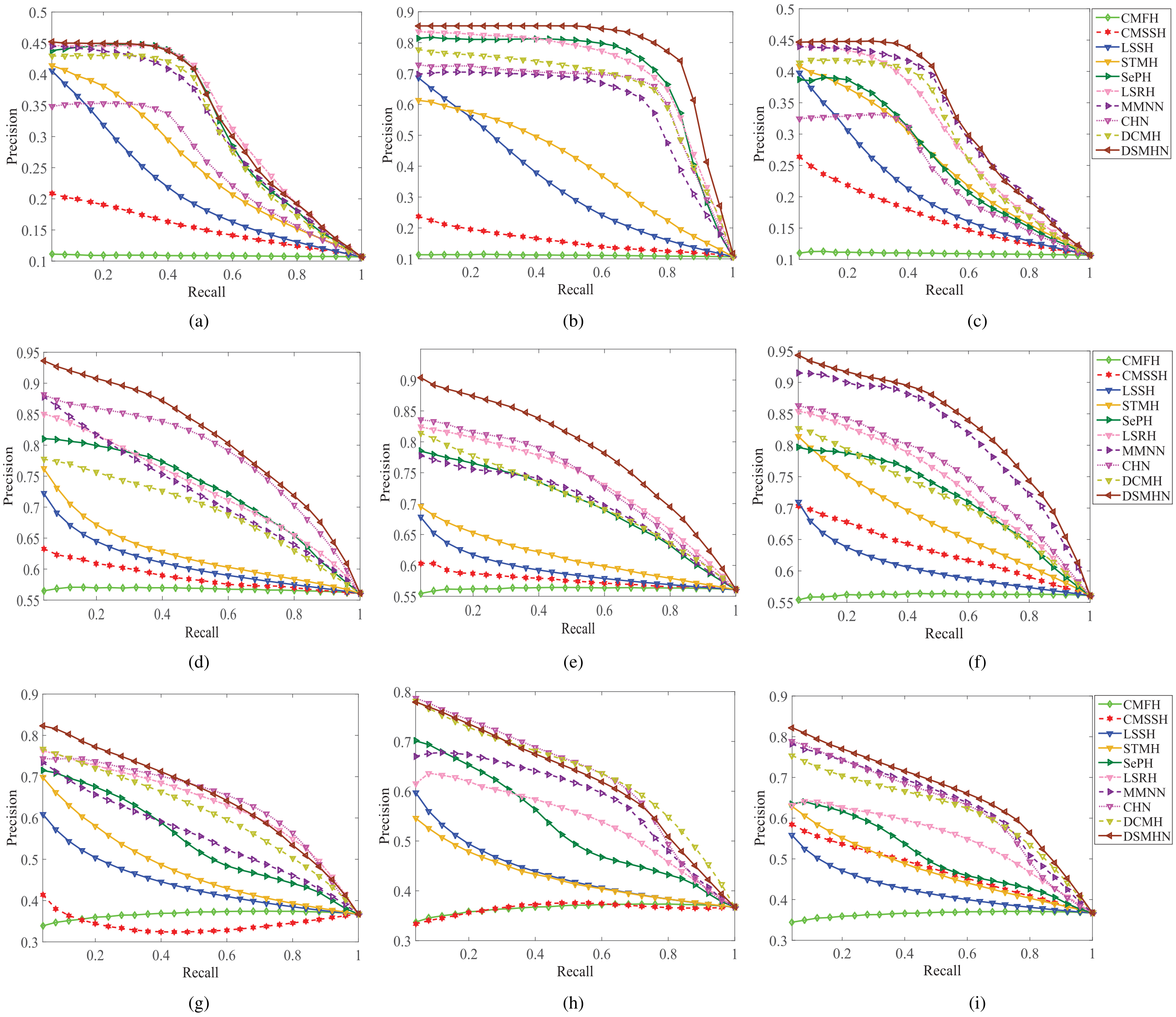

Fig. 4. PR curves for all the methods on Wiki, MIRFlickr25k, and NUS-WIDE with 48 hash bits. (a) Wiki (image-query-text). (b) Wiki (text-query-image). (c) Wiki (image-query-image). (d) MIRFlickr25k (image-query-text). (e) MIRFlickr25k (text-query-image). (f) MIRFlickr25k (image-query-image). (g) NUS-WIDE (image-query-text). (h) NUS-WIDE (text-query-image). (i) NUS-WIDE (image-query-image).

The proposed method is implemented by using the open-source PyTorch framework on an NVIDIA K20 GPU server. The image and video backbones are initialized by coping the parameters from the pretrained AlexNet and I3D models, respectively. The network is trained by using the minibatch SGD with the BP algorithm. We train the network with 350 epochs and set the batch size as 256 and 16 for the image and video backbones, respectively. The learning rate is initialized as $10^{-4}$ and $10^{-2}$ for the backbone and last two fully connected layers, respectively. In the DSMHN, we set $\alpha = 1$ and $\beta = \gamma = 0.5$ for all the data sets. In addition, the contrastive loss is used as the default loss function for the similarity-preserving hash learning.

### D. Experimental Results

For the image-text retrieval, results in terms of mAP cross different code lengths are reported in Table I. In Table I, we compare the retrieval performance of our DSMHN method with six nondeep baselines and three deep baselines. From

TABLE I
COMPARISON IN TERMS OF mAP ON Wiki, MIRFLICKR25K, AND NUS-WIDE WITH 8, 16, 32, AND 48 HASH BITS

| Method | Wiki | | | | MIRFlickr25k | | | | NUS-WIDE | | | |
|---|---|---|---|---|---|---|---|---|---|---|---|---|
| | 8 bits | 16 bits | 32 bits | 48 bits | 8 bits | 16 bits | 32 bits | 48 bits | 8 bits | 16 bits | 32 bits | 48 bits |
| Image-query-text | | | | | | | | | | | | |
| CMFH | 11.50 | 11.69 | 11.92 | 12.12 | 56.54 | 56.66 | 57.03 | 57.22 | 36.48 | 37.24 | 37.38 | 37.67 |
| CMSSH | 22.47 | 24.32 | 21.04 | 20.35 | 60.85 | 60.38 | 58.39 | 59.24 | 47.90 | 42.72 | 41.03 | 40.33 |
| LSSH | 32.61 | 31.63 | 29.91 | 28.64 | 60.50 | 60.95 | 61.30 | 62.00 | 42.20 | 43.77 | 45.24 | 46.97 |
| STMH | 27.06 | 32.47 | 34.52 | 35.66 | 60.58 | 62.63 | 62.92 | 63.35 | 44.79 | 48.20 | 50.69 | 51.12 |
| SePH | 44.57 | 47.52 | 51.08 | 51.48 | 74.61 | 76.33 | 76.84 | 77.07 | 57.56 | 59.25 | 61.23 | 62.48 |
| LSRH | 39.72 | 50.12 | 51.14 | 51.34 | 69.01 | 72.99 | 74.37 | 75.08 | 57.80 | 63.33 | 64.78 | 66.75 |
| MMNN | 38.45 | 41.81 | 44.38 | 46.25 | 71.71 | 72.88 | 74.04 | 74.34 | 55.89 | 56.23 | 58.58 | 58.64 |
| CHN | 37.99 | 38.38 | 38.61 | 40.61 | 72.04 | 73.92 | 75.18 | 80.67 | 59.56 | 63.55 | 67.07 | 67.45 |
| DCMH | 41.34 | 46.21 | 48.67 | 48.72 | 68.65 | 70.19 | 71.58 | 71.55 | 57.18 | 60.34 | 63.50 | 63.76 |
| DSMHN | **48.16** | **51.97** | **52.88** | **53.13** | **80.35** | **82.41** | **83.02** | **84.62** | **63.57** | **67.43** | **67.32** | **69.79** |
| Text-query-image | | | | | | | | | | | | |
| CMFH | 11.53 | 11.59 | 11.82 | 11.99 | 56.80 | 56.93 | 57.26 | 57.39 | 36.63 | 37.12 | 37.40 | 37.34 |
| CMSSH | 22.03 | 20.89 | 16.41 | 16.29 | 58.55 | 59.79 | 58.68 | 56.35 | 41.29 | 37.32 | 36.44 | 35.20 |
| LSSH | 47.19 | 44.58 | 41.55 | 40.35 | 59.81 | 59.91 | 59.93 | 60.04 | 43.29 | 44.42 | 45.33 | 46.34 |
| STMH | 26.91 | 35.46 | 43.83 | 47.55 | 60.03 | 62.20 | 62.26 | 63.10 | 41.05 | 42.73 | 43.17 | 45.23 |
| SePH | 77.83 | 78.10 | 82.79 | 83.48 | 71.99 | 72.76 | 73.36 | 74.10 | 56.18 | 57.89 | 59.08 | 60.11 |
| LSRH | 56.41 | 71.28 | 76.74 | 80.49 | 69.46 | 73.82 | 75.18 | 76.24 | 51.01 | 54.48 | 54.49 | 57.34 |
| MMNN | 50.59 | 55.06 | 61.56 | 69.67 | 72.82 | 73.05 | 73.35 | 75.15 | 57.72 | 58.59 | 60.04 | 62.47 |
| CHN | 68.94 | 68.11 | 71.33 | 73.83 | 73.27 | 74.88 | 75.81 | 77.60 | 60.92 | 65.08 | **66.90** | **68.58** |
| DCMH | 70.07 | 72.65 | 71.52 | 73.30 | 72.84 | 73.89 | 74.43 | 73.60 | 60.10 | 63.56 | 66.19 | 66.84 |
| DSMHN | **83.88** | **85.81** | **86.89** | **88.21** | **80.82** | **81.27** | **82.53** | **82.85** | **61.58** | **66.47** | 66.31 | 68.31 |
| Image-query-image | | | | | | | | | | | | |
| CMFH | 11.41 | 11.59 | 11.59 | 11.92 | 56.41 | 56.56 | 56.86 | 57.07 | 36.50 | 37.01 | 37.23 | 37.14 |
| CMSSH | 19.06 | 18.70 | 18.40 | 20.90 | 64.71 | 64.02 | 65.04 | 65.85 | 47.37 | 45.84 | 48.44 | 47.65 |
| LSSH | 30.93 | 30.04 | 28.83 | 27.65 | 59.91 | 60.34 | 60.68 | 61.38 | 40.25 | 41.65 | 42.90 | 44.21 |
| STMH | 22.01 | 27.02 | 32.17 | 34.81 | 64.92 | 67.30 | 68.02 | 68.40 | 43.83 | 46.28 | 47.92 | 50.39 |
| SePH | 29.77 | 32.65 | 37.25 | 37.43 | 73.41 | 74.95 | 75.52 | 75.90 | 52.36 | 53.68 | 55.38 | 56.29 |
| LSRH | 31.24 | 38.19 | 40.57 | 41.24 | 73.38 | 76.75 | 76.83 | 77.56 | 52.79 | 56.31 | 56.51 | 58.03 |
| MMNN | 46.25 | 44.19 | 48.09 | 48.44 | 84.08 | 83.07 | 84.36 | 85.26 | 65.79 | 65.95 | 67.70 | 67.87 |
| CHN | 40.99 | 39.56 | 39.92 | 41.08 | 78.72 | 80.84 | 81.72 | 78.21 | 65.57 | 63.95 | 66.25 | 67.41 |
| DCMH | 41.91 | 44.78 | 44.97 | 45.67 | 71.93 | 74.33 | 75.29 | 74.25 | 58.79 | 61.45 | 64.87 | 65.20 |
| DSMHN | **50.23** | **53.37** | **53.86** | **53.73** | **83.73** | **84.74** | **86.47** | **87.18** | **66.29** | **68.99** | **67.98** | **70.70** |

Table I, we can observe that our DSMHN method consistently outperforms the compared baselines across different data sets and code lengths. Specifically, compared with SePH and LSRH (the state-of-the-art nondeep baselines), the average performance increases of our DSMHN method are 2.85%/5.64%/14.98%, 6.38%/7.56%/9.39%, and 3.86%/7.35%/12.62% for image-query-text/text-query-image/image-query-image on the Wiki, MIRFlickr25k, and NUS-WIDE data sets, respectively. Similarly, when compared with the best results of deep baselines, a DSMHN gains the average performance increases of 5.3%/17.65%/6.06%, 7.14%/6.47%/1.34%, and 2.62%/0.67%/1.66% on the Wiki, MIRFlickr25k, and NUS-WIDE data sets, respectively. In addition, we find that the performance gap between different retrieval tasks is very small for the MIRFlickr25k and NUS-WIDE data sets except for the Wiki data set. This is mainly caused by the huge semantic gap between the image and text modalities in the Wiki data set. As the text in the Wiki data set is much better than the image in describing the semantic concepts, much better performance can be achieved when the text is used to query the image in the database. Another interesting observation is that the mAP results of the DSMHN for the image-query-image task are much better than the compared baselines across different data sets. In the DSMHN, the intramodality semantic class labels are explicitly exploited in hashing learning; thus, the intramodality correlation is maximally preserved. This makes the learned hash codes discriminative to achieve superior performance for the image-retrieval task.

We also evaluate the retrieval performance in terms of P@100 and PR. Fig. 2 shows the performance in terms of P@100 across different code bits and data sets. Fig. 3 shows the precision curves with respect to different numbers of top returned samples for 48-bit hash codes across different data sets. From Figs. 2 and 3, we can observe that a DSMHN substantially outperforms all the baselines for both single-modal- and cross-modal-retrieval tasks. Fig. 4 shows the retrieval performance in terms of PR with 48 hash bits across different data sets. Note that the larger area under the PR curve indicates better overall performance. As shown in Fig. 4, the DSMHN is competitive or outperforms all the baselines across different data sets.

For the video-text retrieval, we adopt a 3-D CNN, such as I3D [77] network, as the video backbone to capture the spatial and temporal information. The results in terms of mAP and P@100 of all methods are presented in Table II. We can see that the mAP values are much larger than all the compared deep baselines across different retrieval tasks and code lengths. Specifically, compared with the state-of-the-art baseline MMNN, the DSMHN achieves the mAP increases of

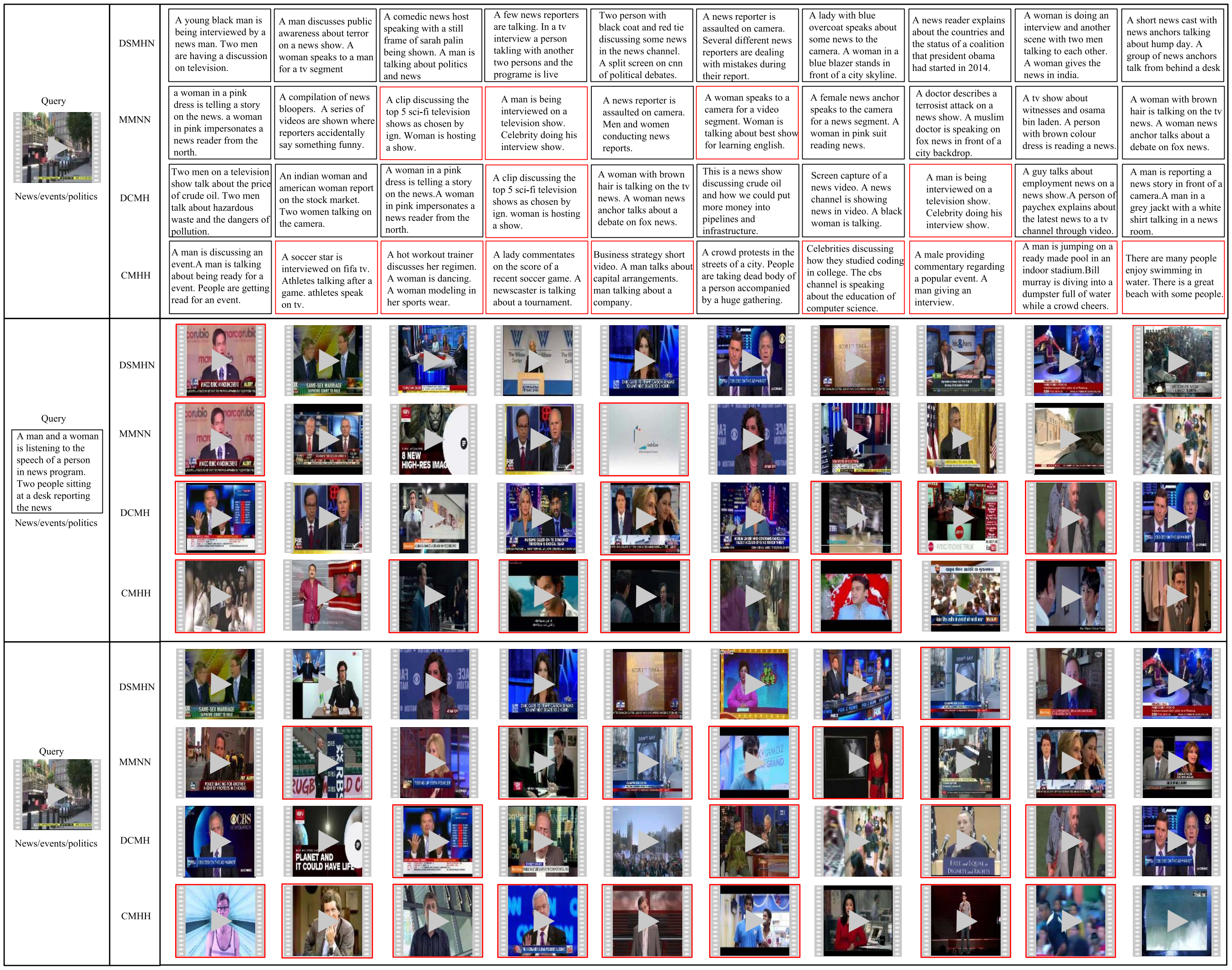

Fig. 5. Examples of top ten returned samples on MSR-VTT-10K with 48 hash bits. From top to bottom, it shows the examples for the video-query-text, text-query-video, and video-query-video task, respectively. Results are returned based on the Hamming ranking, and approximately two of the most relevant annotations are shown. The red rectangles indicate the wrong results.

TABLE II

COMPARISON IN TERMS OF mAP AND P@100 ON MSR-VTT-10K WITH 8, 16, 32, AND 48 HASH BITS

| Method | Video-query-text | | | | Text-query-video | | | | Video-query-video | | | |
|---|---|---|---|---|---|---|---|---|---|---|---|---|
| | 8 bits | 16 bits | 32 bits | 48 bits | 8 bits | 16 bits | 32 bits | 48 bits | 8 bits | 16 bits | 32 bits | 48 bits |
| mAP | | | | | | | | | | | | |
| MMNN | 28.21 | 31.47 | 33.11 | 33.67 | 24.67 | 26.19 | 26.29 | 26.75 | 19.52 | 21.39 | 21.86 | 22.44 |
| DCMH | 25.08 | 28.36 | 29.32 | 31.04 | 18.67 | 23.61 | 25.72 | 25.40 | 16.21 | 17.30 | 19.85 | 20.31 |
| CMHH | 25.56 | 27.13 | 29.16 | 29.69 | 20.21 | 23.33 | 24.13 | 32.43 | 16.43 | 19.01 | 20.01 | 20.31 |
| DSMHN | **32.35** | **35.19** | **37.39** | **37.92** | **31.09** | **34.23** | **37.32** | **37.35** | **19.54** | **23.41** | **26.00** | **26.91** |
| P@100 | | | | | | | | | | | | |
| MMNN | **29.15** | 32.31 | 34.38 | 34.35 | 34.60 | 35.30 | 35.88 | 36.08 | **24.76** | 27.06 | 27.97 | 28.72 |
| DCMH | 24.80 | 28.77 | 30.82 | 33.23 | 23.52 | 34.07 | 35.61 | 34.81 | 18.94 | 23.28 | 25.90 | 26.69 |
| CMHH | 25.80 | 27.30 | 29.91 | 29.69 | 26.37 | 30.80 | 32.61 | 32.43 | 19.03 | 22.86 | 24.64 | 25.45 |
| DSMHN | 28.66 | **33.42** | **34.50** | **34.96** | **41.56** | **44.56** | **47.94** | **47.22** | 23.65 | **28.24** | **31.13** | **31.80** |

4.10%, 9.02%, and 2.66% on average for the video-query-text, text-query-video, and video-query-video tasks, respectively. The DSMHN is competitive to or outperforms the compared baselines across different retrieval tasks and code lengths in terms of P@100. In general, the significant improvement shows the proposed method with the advantage in the ability of capturing more useful spatial and temporal information to learn effective hash codes for both single-modal and cross-modal retrievals.

The retrieval results on MSR-VTT-10K for video-query-text, text-query-video, and video-query-video are shown in Fig. 5. For each given query, we return top ten results based on the Hamming ranking and show them in the right column of Fig. 5. The wrong results are marked with red rectangles. It can be seen that our DSMHN method can return much more relevant samples than the compared deep methods. The impressive results demonstrate that our DSMHN method can generate efficient hash codes for both single-modal and cross-modal retrievals by jointly exploiting the intermodality correlation structure and the intramodality semantic information.

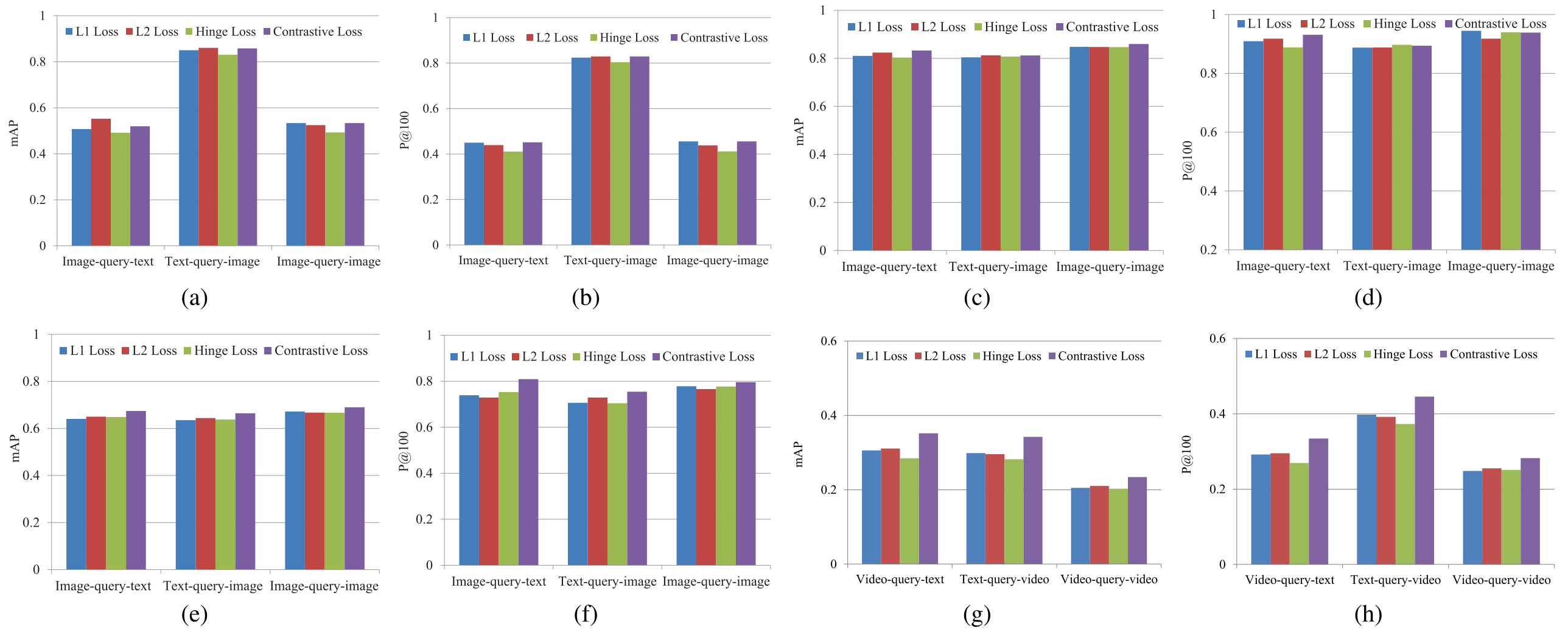

Fig. 6. Compared results of DSMHN with different loss functions on Wiki, MIRFlickr25k, NUS-WIDE, and MSR-VTT-10K with 16 hash bits. (a) Wiki (mAP). (b) Wiki (P@100). (c) MIRFlickr25k (mAP). (d) MIRFlickr25k (P@100). (e) NUS-WIDE (mAP). (f) NUS-WIDE (P@100). (g) MSR-VTT-10K (mAP). (h) MSR-VTT-10K (P@100).

### *E. Discussion of Different Loss Functions*

The proposed deep hashing method can be integrated with different types of loss functions. In this section, we evaluate the performance of the proposed DSMHN according to mAP and P@100 by using four different types of loss functions such as L1 loss, L2 loss, hinge loss, and contrastive loss. The details of these loss functions can be found in Section III-B. We report the mAP and P@100 results with a 16-bit hash code for four different loss functions in Fig. 6. We observe that the performance of different loss functions is very close. Such a small performance difference demonstrates the robustness of the proposed method with different loss functions. In addition, the capability of integrating different loss functions into the unified deep hashing framework shows the scalability and flexibility of the proposed DSMHN method.

### *F. Hyperparameter Sensitivity Analysis*

We conduct the sensitivity analysis of the hyperparameters $\alpha$, $\beta$, and $\gamma$ on the MSR-VTT-10K data set. We perform the linear search in the set $\{0, 0.0001, 0.01, 0.1, 0.5, 1\}$ and select the best hyperparameters for our DSMHN method. We analyze the impact of one hyperparameter with the others fixed to the default values. The results in terms of mAP and P@100 are shown in Fig. 7, where the hash bit is set to 16. From Fig. 7, we can clearly see that our DSMHN can achieve the best performance when $\alpha$ is set to 1. Moreover, the performance is stable when $\beta$ and $\gamma$ are in $[0.5, 1]$ and $[0.01, 0.5]$, respectively. Overall, we set $\alpha = 1$ and $\beta = \gamma = 0.5$ for all the data sets.

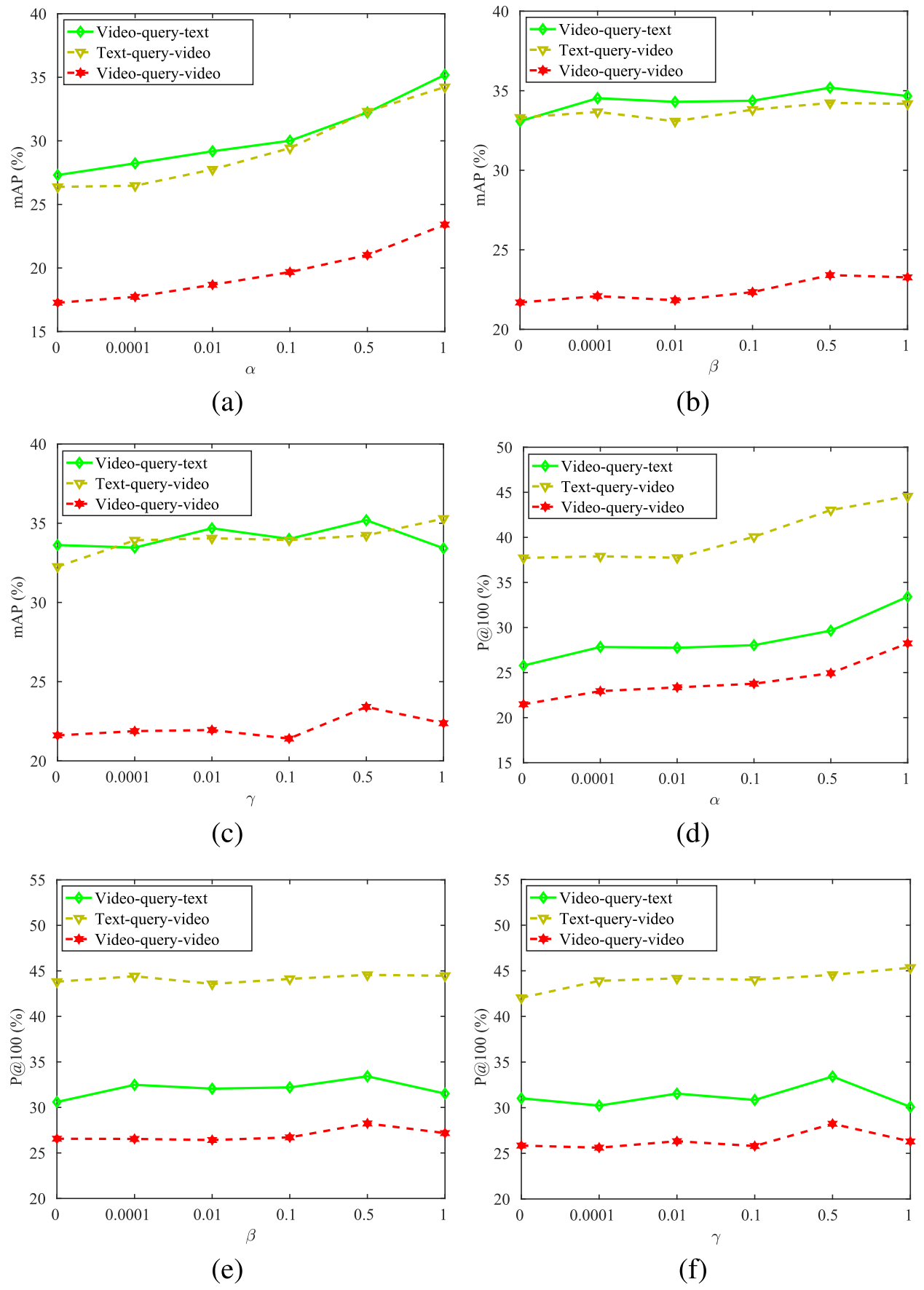

Fig. 7. Hyperparameter sensitivity analysis on the MSR-VTT-10K data set with 16 hash bits. (a) $\alpha$ (mAP). (b) $\beta$ (mAP). (c) $\gamma$ (mAP). (d) $\alpha$ (P@100). (e) $\beta$ (P@100). (f) $\gamma$ (P@100).

## V. CONCLUSION

In this article, we propose a novel DSMHN for scalable image-text and video-text retrievals by exploring the intermodality correlation structure and the intramodality semantic label information with the deep neural networks. Specifically, the DSMHN integrates the feature representation learning, intermodality similarity-preserving learning, intramodality semantic label learning, and hash learning into an end-to-end framework. In addition, the DSMHN can recommend different types of loss functions with minimal modification to the hash layer of the network, which demonstrates the scalability and flexibility of the proposed framework. We conduct extensive experiments on four widely used multimodal data sets for both single-modal and cross-modal-retrieval tasks. The impressive experimental results demonstrate that the DSMHN can generate efficient hash codes, thus achieving the state-of-the-art performance for both image-text and video-text retrievals.

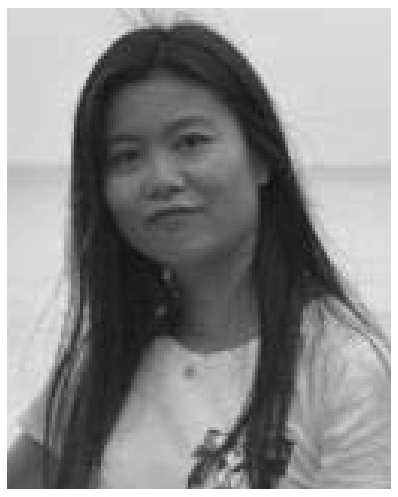

**Lu Jin** received the B.E. degree from Northeast University, Qinhuangdao, China, in 2010, and the Ph.D. degree from the Nanjing University of Science and Technology, Nanjing, China, in 2019.

Her research interests include multimedia computing, deep learning, and multimedia retrieval.

Dr. Jin has received the Best Student Paper Award in ICIMCS 2018.

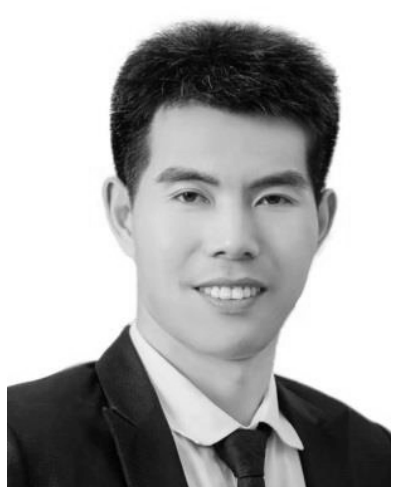

**Zechao Li** (Member, IEEE) received the B.E. degree from the University of Science and Technology of China, Hefei, China, in 2008, and the Ph.D. degree from the National Laboratory of Pattern Recognition, Institute of Automation, Chinese Academy of Sciences, Beijing, China, in 2013.

He is currently a Professor with the Nanjing University of Science and Technology, Nanjing, China. He has authored more than 60 articles in top-tier journals and conferences. His research interests include big media analysis, computer vision, and so on.

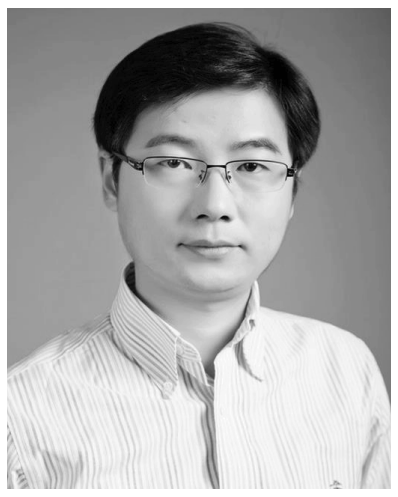

**Jinhui Tang** (Senior Member, IEEE) received the B.E. and Ph.D. degrees from the University of Science and Technology of China, Hefei, China, in July 2003 and July 2008, respectively.

From 2008 to 2010, he was a Research Fellow with the School of Computing, National University of Singapore, Singapore. He is currently a Professor with the School of Computer Science and Engineering, Nanjing University of Science and Technology, Nanjing, China. His current research interests include multimedia computing and computer vision. He has authored over 100 journal and conference papers in these areas.

Dr. Tang was a co-recipient of the best paper awards in ACM MM 2007, PCM 2011, and ICIMCS 2011, the Best Paper Runner-Up in ACM MM 2015, and the Best Student Paper Awards in MMM 2016 and ICIMCS 2017. He has served as an Associate Editor for the IEEE TRANSACTIONS ON NEURAL NETWORKS AND LEARNING SYSTEMS, the IEEE TRANSACTIONS ON MULTIMEDIA, the IEEE TRANSACTIONS ON KNOWLEDGE AND DATA ENGINEERING, and the IEEE TRANSACTIONS ON CIRCUITS AND SYSTEMS FOR VIDEO TECHNOLOGY.